\documentclass[10pt,twocolumn,letterpaper]{article}

\usepackage{cvpr}
\makeatletter
\@namedef
{ver@everyshi.sty}{}
\makeatother
\usepackage{cmap}
\usepackage{times}
\usepackage{epsfig}
\usepackage{graphicx}
\usepackage{amsmath}
\usepackage{amssymb}
\usepackage[dvipsnames]{xcolor}
\usepackage{bm}
\usepackage{xspace}
\usepackage{subcaption}
\usepackage{adjustbox}
\usepackage{booktabs}
\usepackage{diagbox}
\usepackage{multirow}
\usepackage[inline]{enumitem}
\usepackage[linesnumbered,ruled]{algorithm2e}
\usepackage[pagebackref=true,breaklinks=true,letterpaper=true,colorlinks,bookmarks=false, citecolor=cyan, linkcolor=magenta]{hyperref}

\cvprfinalcopy % *** Uncomment this line for the final submission

\def\cvprPaperID{1} % *** Enter the CVPR Paper ID here

\begin{document}
	
	\newcommand{\im}{I}
	\newcommand{\tplgen}{G}
	\newcommand{\backboneblock}{\tplgen_{B}}
	\newcommand{\attblock}{\tplgen_{A}}
	\newcommand{\fuseblock}{\tplgen_{F}}
	\newcommand{\att}{A}
	\newcommand{\attr}{a}
	\newcommand{\simscore}{s}
	\newcommand{\simmtx}{S}
	
	\newcommand{\tpl}{t}
	\newcommand{\globaltpl}{\tpl^{g}}
	\newcommand{\localtpl}{\tpl^{l}}
	\newcommand{\fusetpl}{\tpl^{f}}
	
	\newcommand{\gtlabel}{y}
	\newcommand{\predlabel}{\hat{\gtlabel}}
	\newcommand{\bs}{M}
	\newcommand{\margin}{m}
	\newcommand{\oreo}{OREO\xspace}
	%%%%%%%%% TITLE
	\title{On Improving the Generalization of Face Recognition \\ in the Presence of Occlusions}
	
	\author{Xiang Xu, Nikolaos Sarafianos, and Ioannis A. Kakadiaris\\
		Computational Biomedicine Lab\\
		University of Houston\\
		{\tt\small \{xxu21, nsarafianos, ioannisk\}@uh.edu}
	}
	
	\maketitle
	%\thispagestyle{empty}
	
	%%%%%%%%% ABSTRACT
	\begin{abstract}
		In this paper, we address a key limitation of existing 2D face recognition methods: robustness to occlusions. To accomplish this task, we systematically analyzed the impact of facial attributes on the performance of a state-of-the-art face recognition method and through extensive experimentation, quantitatively analyzed the performance degradation under different types of occlusion.
		Our proposed Occlusion-aware face REcOgnition (OREO) approach learned discriminative facial templates despite the presence of such occlusions.
		First, an attention mechanism was proposed that extracted local identity-related region. The local features were then aggregated with the global representations to form a single template.
		Second, a simple, yet effective, training strategy was introduced to balance the non-occluded and occluded facial images.
		Extensive experiments demonstrated that \oreo improved the generalization ability of face recognition under occlusions by \(10.17\%\) in a single-image-based setting and outperformed the baseline by approximately \(2\%\) in terms of rank-1 accuracy in an image-set-based scenario. 
	\end{abstract}
	
	\section{Introduction}
	The goal of this paper is to present a face recognition method that is robust to facial occlusions originating from facial attributes. 
	For example, given a facial image of an individual wearing sunglasses or a hat, we aspire to successfully match this probe image with the corresponding images in the gallery to obtain his/her identity. Note that there are other challenges (such as self-occlusions or extreme pose variations) that might affect the face recognition performance. External occlusions can be defined as those caused by facial accessories such as glasses, hats or different types of facial hair.
	Despite the recent success of face recognition methods~\cite{Deng_2019_180917, Schroff_2015_180917, Taigman_2014_180921}, most existing research tends to focus solely on the pose challenge while failing to account for other factors such as occlusion, age, and facial expression variations, that can have a negative impact on the face recognition performance.
	
	\begin{figure}[tb]
		\centering
		\begin{subfigure}[b]{0.34\linewidth}
			\centering
			\includegraphics[height=2.5in]{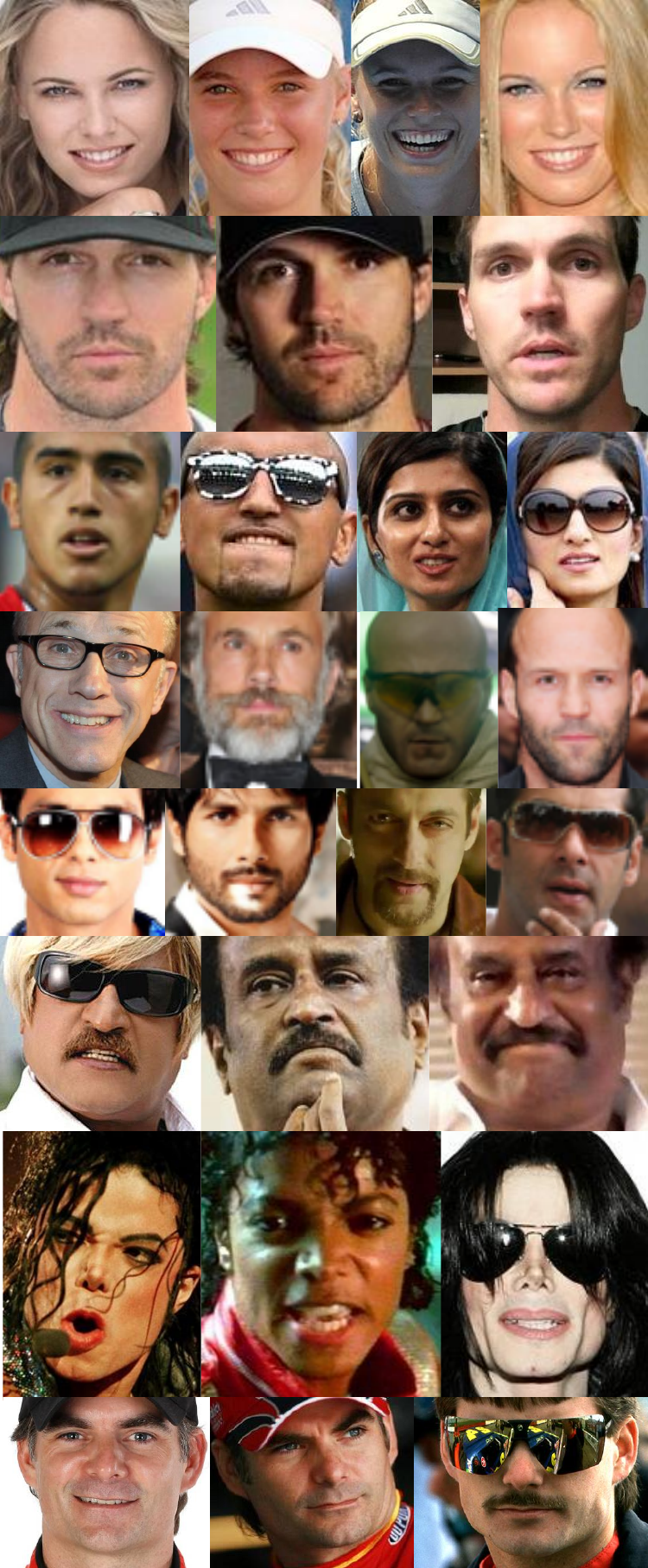}
			\caption{}
		\end{subfigure}
		\begin{subfigure}[b]{0.31\linewidth}
			\centering
			\includegraphics[height=2.5in]{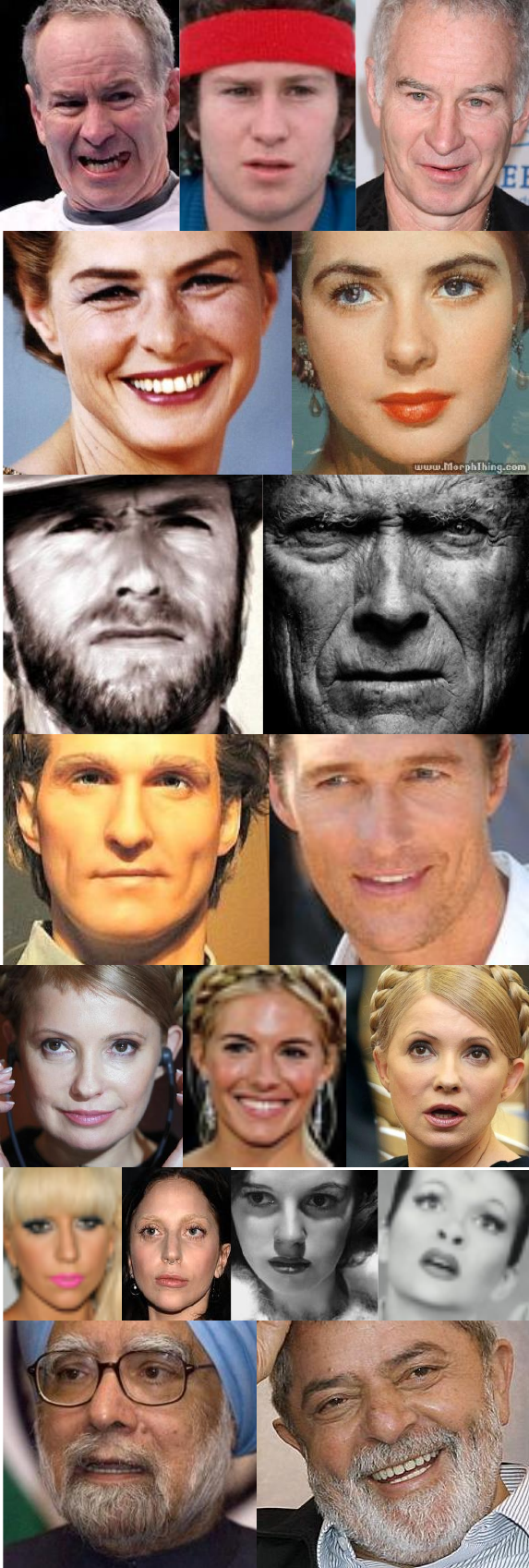}
			\caption{}
		\end{subfigure}
		\begin{subfigure}[b]{0.31\linewidth}
			\centering
			\includegraphics[height=2.5in]{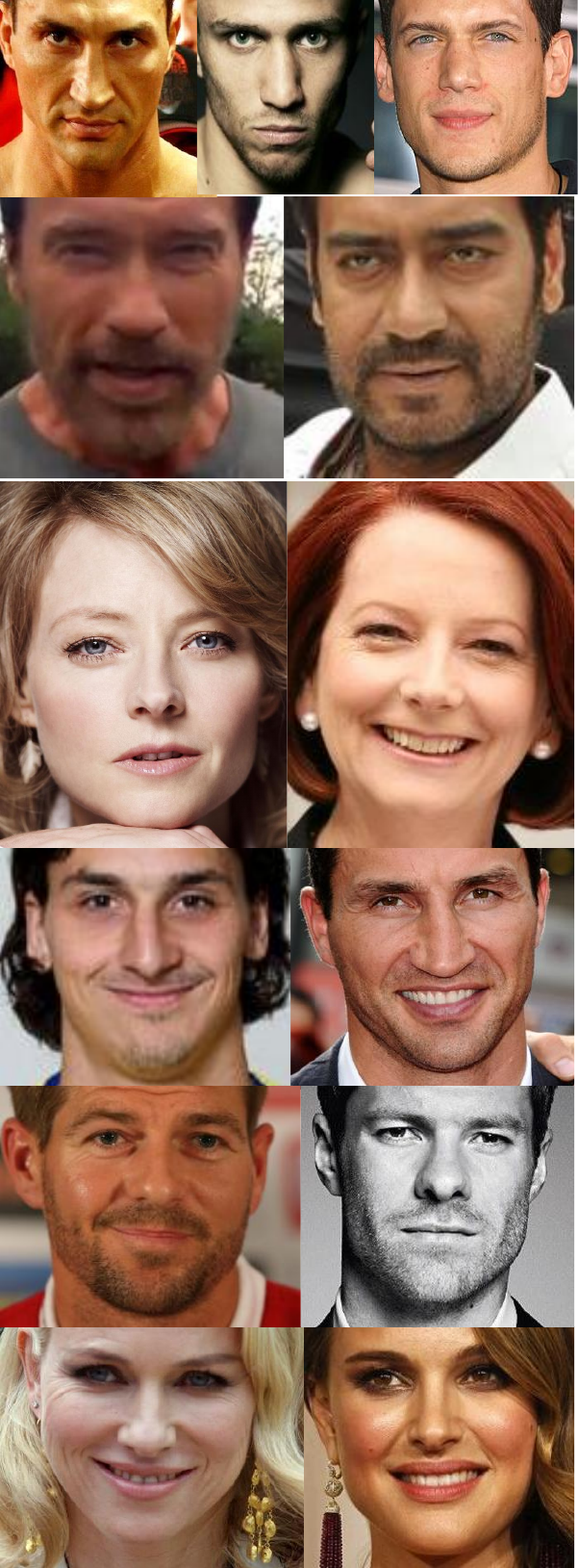}
			\caption{}
		\end{subfigure}
		\caption{Depiction of incorrectly matched samples using ResNeXt-101 on the CFP-FF dataset: (a) false negative matches due to occlusions; (b) false negative matches due to age, pose, and facial expression variations; and (c) false positive matches due to similar appearance.}
		\vspace{-1.5em}
		\label{FIG:DRFR-CFP-Sample}
	\end{figure}
	
	Aiming to gain a better understanding of the common failure cases, a ResNeXt-101~\cite{He_2016_190212} model was trained on the VGGFace2~\cite{Cao_2018_17830} dataset and was evaluated on the CFP dataset~\cite{Sengupta_2016_17835} using the frontal-to-frontal matching protocol (CFP-FF). 
	This model was selected to serve as a baseline since its verification performance on the CFP dataset is almost state-of-the-art and at the same time, it is easy to train in all deep learning frameworks. 
	Figure~\ref{FIG:DRFR-CFP-Sample} presents the false positive and false negative pairs of images from the CFP dataset based on predictions of the ResNeXt-101 model.
	It is worth noting that in this protocol the faces have low variation in the yaw angle. 
	The obtained results indicate that the sources of error for most false matching results originate from factors such as occlusion and age difference. 
	Similar results are observed in the same dataset using the frontal-to-profile matching protocol (CFP-FP).
	Based on these observations, we can confidently conclude that besides large pose variations, occlusion is a significant factor that greatly affects the recognition performance. 
	
	Why does face recognition performance degrade in the presence of occlusions? First, important identity-related information might be excluded when the face is occluded. Figure~\ref{FIG:DRFR-CFP-Sample}(a) depicts several samples that would be challenging for a human to decide whether the individuals with sunglasses belong to the same identity or not.
	Second, existing deep learning approaches are data-driven, which means that the generalization power of the model usually is limited by the training data. 
	However, current datasets are collected by focusing mainly on the pose distribution. 
	This introduces a large class imbalance in terms of occlusion since, in the majority of the selected samples, the entire face is visible. 
	A limited number of approaches~\cite{Cheng_2015_180928, He_2018_180926, Zhao_2018_180928} have recently tried to address this problem.
	However, such methods require prior knowledge of occlusion presence~\cite{Cheng_2015_180928, Zhao_2018_180928} to perform de-occlusion on synthetically generated occlusions, which is not practical in real-life applications.
	
	Aiming to improve the generalization of face recognition methods, we first investigate which type of occlusion affects the most the final performance. Our experimental analysis indicated that the main reasons for the performance degradation in the presence of occlusions are: (i) identity signal degradation (\ie, information related to the identity of the individual is lost in the presence of occlusion), and (ii) occlusion imbalance in the training set.  
	To address the first challenge of identity signal degradation due to occlusions, an attention mechanism is introduced which is learned directly from the training data. 
	Since the global facial template captures information learned from the whole facial image (regardless of whether occlusion occurs), the attention mechanism aims to disentangle the identity information using the global representation and extract local identity-related features from the non-occluded regions of the image. 
	In this way, global and local features are jointly learned from the facial images and are then aggregated into a single template.
	To address the challenge of the occlusion imbalance in the training set, an occlusion-balanced sampling strategy is designed to train the model with batches that are equally balanced with non-occluded and occluded images. 
	Based on this strategy, an additional learning objective is proposed that improves the discriminative ability of the embeddings learned from our algorithm. 
	Thus, the proposed occlusion-aware feature generator results in facial embeddings that are robust to occlusions originating from visual attributes. Our results demonstrate that \oreo significantly improves the face recognition performance without requiring any prior information or additional supervision.
	Through extensive experiments and ablation studies, we demonstrate that the proposed approach achieves comparable or better face recognition performance on non-occluded facial images and at the same time significantly improves the generalization ability of the facial embedding generator on facial images in which occlusion is present.

	In summary, the contributions of this work are:
	\begin{enumerate*}[label=(\roman*)]
		\item An analysis of the impact of attributes to the face recognition performance is conducted on the Celeb-A dataset and its key insights are presented.
		\item An attention mechanism is introduced that disentangles the features into global and local parts, resulting in more discriminative representations. 
		In this way, the global features contain identity-related information while the local features learned through our attention mechanism are robust to occlusions caused by visual attributes. 
		\item An occlusion-balanced sampling strategy along with a new loss function are proposed to alleviate the large class imbalance that is prevalent due to non-occluded images in existing datasets.
	\end{enumerate*} 
	
	\begin{figure*}[htb]
		\centering
		\begin{subfigure}[b]{0.33\linewidth}
			\centering
			\includegraphics[width=\linewidth]{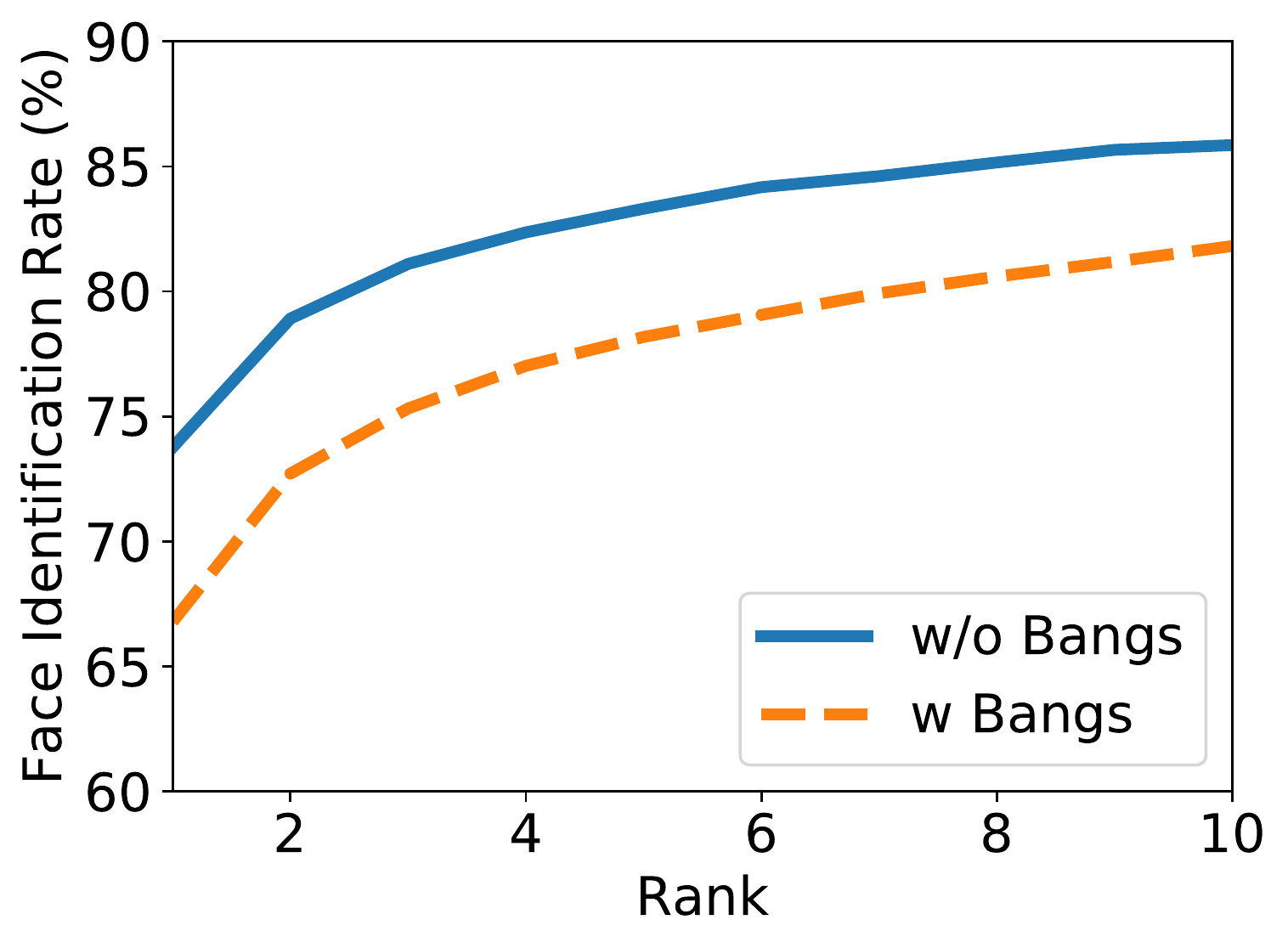}
			\vspace{-2em}
			\caption{}
		\end{subfigure}
		\begin{subfigure}[b]{0.33\linewidth}
			\centering
			\includegraphics[width=\linewidth]{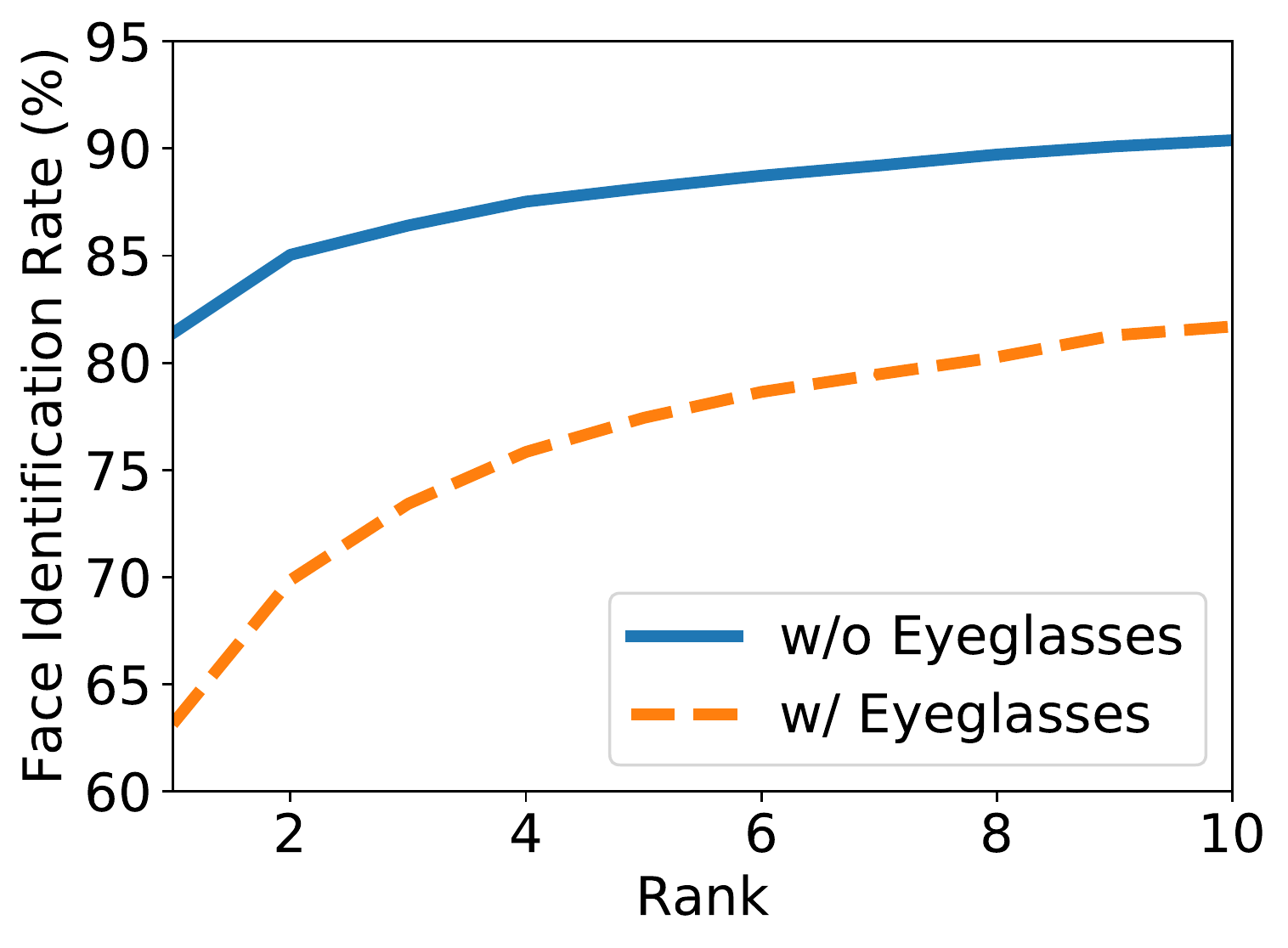}
			\vspace{-2em}
			\caption{}
		\end{subfigure}
		\begin{subfigure}[b]{0.33\linewidth}
			\centering
			\includegraphics[width=\linewidth]{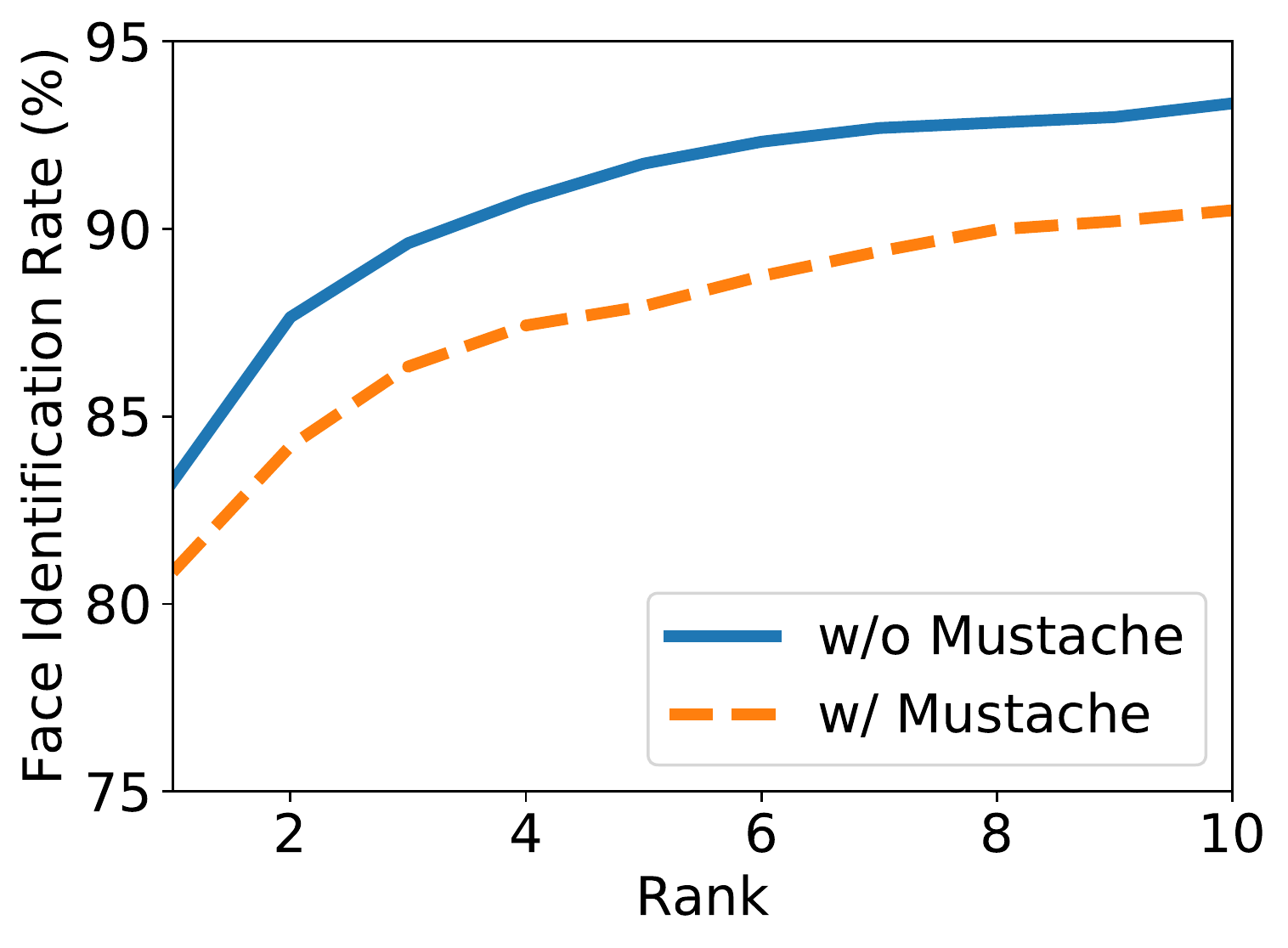}
			\vspace{-2em}
			\caption{}
		\end{subfigure}
		\begin{subfigure}[b]{0.33\linewidth}
			\centering
			\includegraphics[width=\linewidth]{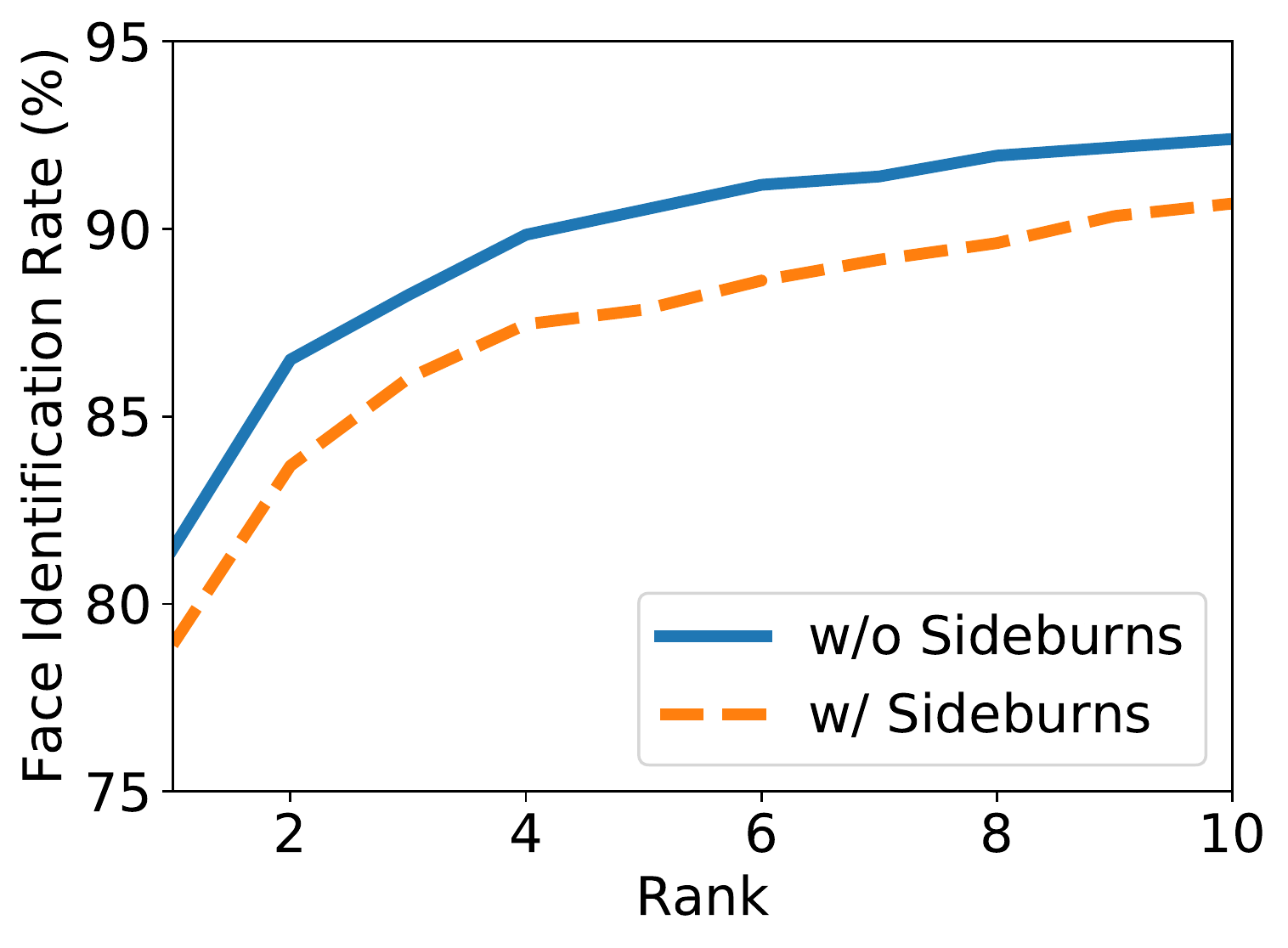}
			\vspace{-2em}
			\caption{}
		\end{subfigure}
		\begin{subfigure}[b]{0.33\linewidth}
			\centering
			\includegraphics[width=\linewidth]{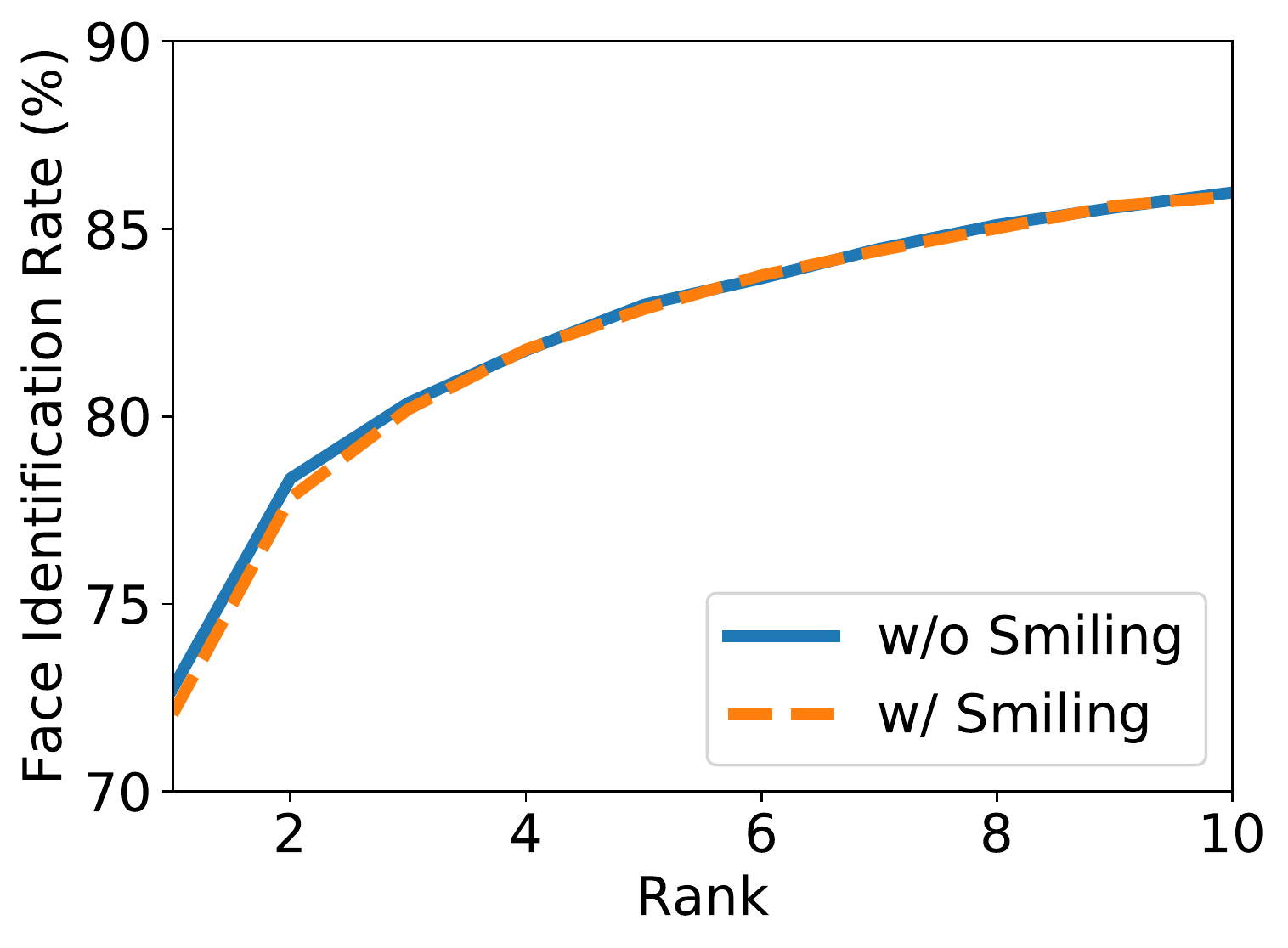}
			\vspace{-2em}
			\caption{}
		\end{subfigure}
		\begin{subfigure}[b]{0.33\linewidth}
			\centering
			\includegraphics[width=\linewidth]{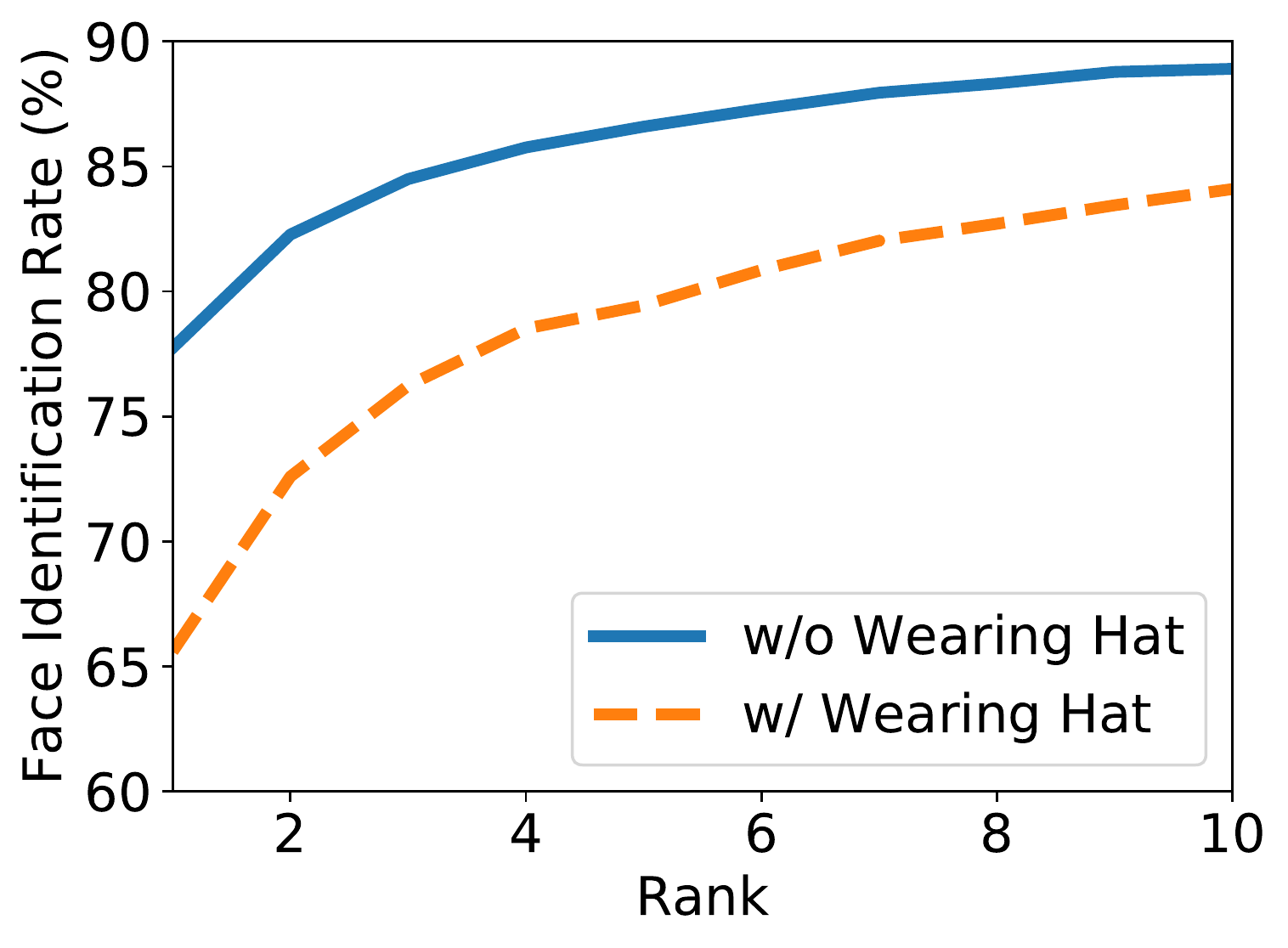}
			\vspace{-2em}
			\caption{}
		\end{subfigure}
		\vspace{-1.5em}
		\caption{Depiction of CMCs demonstrating the impact of selected facial attributes on the face identification performance using a ResNeXt-101 as a backbone architecture that generates facial embeddings. The corresponding Celeb-A attributes are: (a) \texttt{Bangs}, (b) \texttt{Eyeglasses}, (c) \texttt{Mustache}, (d) \texttt{Sideburns}, (e) \texttt{Smiling}, and (f) \texttt{Wearing Hat}.}
		\vspace{-1.5em}
		\label{FIG:DRFR-CelebA-CMC}
	\end{figure*}
	
	\section{Related Work}
	\label{SEC:DRFR-LR}
	\noindent \textbf{Face Recognition under Occlusion}:
	Face recognition techniques that generate 2D frontal images or facial embeddings from a single image have been proposed that: (i) use a 3D model~\cite{Masi_2016_17266, Xu_2017_17643, xu2019importance}, (ii) generative adversarial networks~\cite{Bao_2018_180928, chen2019r3, Deng_2018_181107, Huang_2017_17458, Tran_2017_17705, Yin_2017_17457, Zhao_2018_181005, Zhao_2018_180922}, and (iii) various transformations~\cite{Cao_2018_180927, Zhong_2017_181107, Zhou_2018_180923}. Recent works have also focused on long-tail~\cite{yin2019feature, zhong2019unequal} or noisy data~\cite{hu2019noise}
	Additionally, multiple loss functions~\cite{Deng_2019_180917, duan2019uniformface, fair2019iccv, liu2019adaptiveface, Liu_2017_180923, Wang_2018_181025, Wang_2018_181031, Wen_2016_181003, zhao2019multi, zhao2019regularface, Zheng_2018_181107} have been developed to guide the network to learn more discriminative face representations, but usually, ignore facial occlusions.
	Early methods approached face recognition in the presence of occlusions by using variations of sparse coding~\cite{Fu_2017_180928, Wagner_2012_180928, Yu_2017_180928, Zhao_2015_180928}. 
	However, such techniques work well only with a limited number of identities, and with frontal facial images in a lab controlled environment.
	The works of He \etal~\cite{He_2018_180926} and Wang \etal~\cite{Wang_2016_180928} addressed this limitation by matching face patches \cite{Xu_2017_17643} under the assumption that the occlusion masks were known beforehand and that the occluded faces from the gallery/probe were also known, which is not realistic.
	Guo \etal~\cite{Guo_2018_180927} fit a 3D model on images in-the-wild to render black glasses and enlarge the training set. However, this method was designed to tackle a specific type of occlusion and cannot cover most occlusion cases that might appear in scenarios in the wild. 
	Finally, Song \etal~\cite{song2019occlusion} proposed a mask-learning method to tackle occlusions in face recognition applications. A pairwise differential siamese network was introduced so that correspondences could be built between occluded facial blocks and corrupted feature elements. 
	
	\noindent \textbf{Visual Attention}: 
	Several works have appeared recently that demonstrate the ability of visual attention to learn discriminative feature representations~\cite{Hu_2018_181102, Li_2018_181102, Sarafianos_2018_180921, Wang_2017_181015, Zhu_2017_181102, sarafianos2019adversarial}.
	Most methods~\cite{Chu_2017_181102, Rodriguez_2018_181102, Wang_2017_181102} extract saliency heatmaps at multiple-scales to build richer representations but fail to account for the correlation between the attention masks at different levels. 
	Jetley \etal~\cite{Jetley_2018_181010} proposed a self-attention mechanism that focuses on different regions to obtain local features for image classification under the hypothesis that the global representation contains the class information. 
	However, in their architecture, the global features are not regularized by the identity loss function, which does not support their original hypothesis.
	Castanon \etal~\cite{Castanon_2018_181102} used visual attention to quantify the discriminative region on facial images.
	Other methods~\cite{Ranjan_2018_181003, Xie_2018_181003, Xie_2018_181002, Yang_2017_181102} apply attention mechanisms to weigh the representations from multiple images. Finally, Shi \etal~\cite{Shi_2018_181003} introduced a spatial transformer to find the discriminative region but their method required additional networks to obtain the local features.
	
	In this paper, instead of simulating each case of facial occlusion, a new method is presented that directly learns from images in-the-wild that contain a plethora of occlusion variations. Our approach improves the generalization ability of the facial embedding generator, without having any prior knowledge of whether the occlusion is present.
	
	\begin{figure*}[thb]
		\centering
		\includegraphics[width=\linewidth]{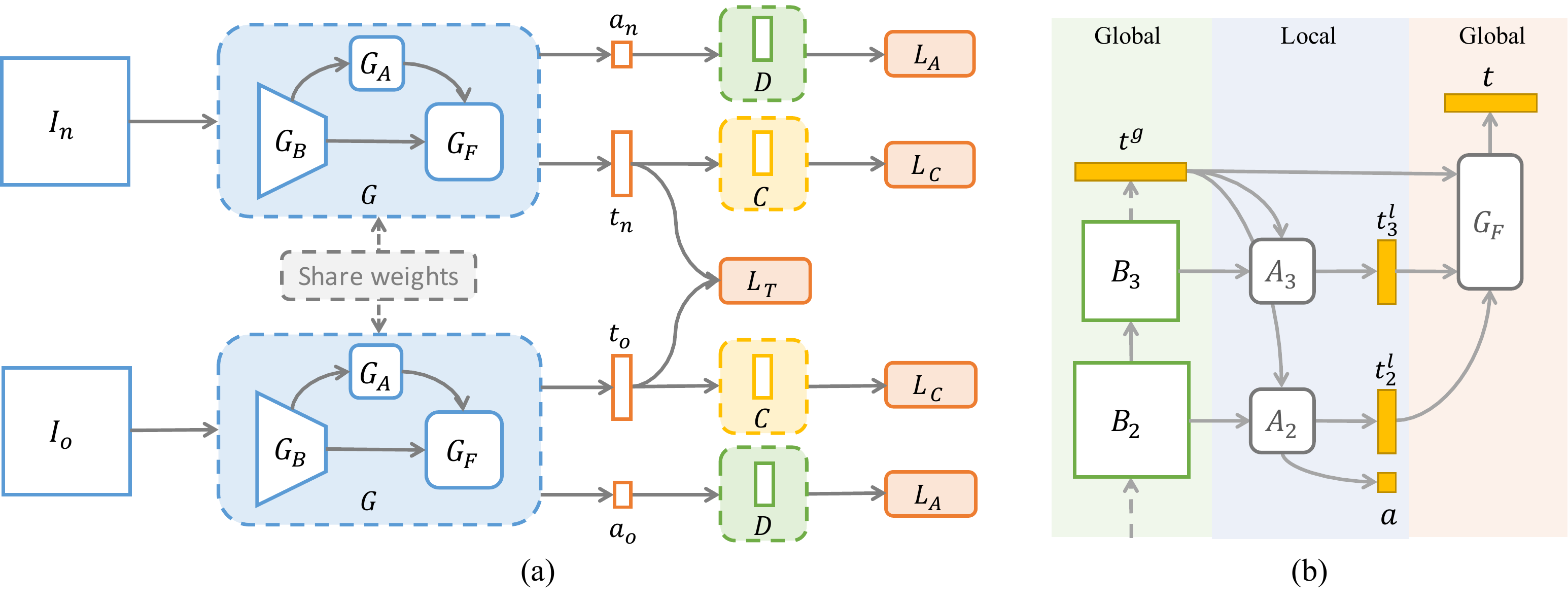}
		\vspace{-2em}
		\caption{(a) Given a pair of non-occluded and occluded images (\(I_n, I_o\)), the template generator \(G\) learns the facial embeddings \((t_n, t_o)\) and the attributes predictions \(a_n, a_o\) with the attributes classifiers ($D$) and identity classifier ($C$) using facial attribute classification loss $L_A$, identity classification loss $L_C$, and the proposed similarity triplet loss $L_T$.
			(b) Depiction of the generator in detail, which contains: (i) the output feature maps of the last two blocks \((B_2, B_3)\) of the backbone architecture, (ii) the attention mechanism \(G_A\) consisting of masks \((A_2, A_3)\) that learn the local features in two different ways, and (iii) \(G_F\) which aggregates the global and local features to the final embedding.}
		\vspace{-1em}
		\label{FIG:DRFR-Framework}
	\end{figure*}
	
	\section{Systematic Analysis on Impact of Facial Attributes on Face Recognition}
	\label{SEC:DRFR-CelebA}
	Aiming to quantitatively analyze the impact of occlusion, a series of experiments are conducted on the Celeb-A dataset~\cite{Liu_2015_180921}, which consists of $10,177$ face identities and $40$ facial attributes.
	Attributes that describe the subject (\eg, \texttt{Gender}) were ignored and only those that might impact the face recognition performance were selected: \texttt{Bangs}, \texttt{Eyeglasses}, \texttt{Mustache}, \texttt{Sideburns}, \texttt{Smiling}, and \texttt{Wearing Hat}. 
	For each attribute, the images without this attribute were enrolled as the gallery and images w/ or w/o this attribute were enrolled as probes.
	In both the gallery and probe, each identity has only a single image.
	A ResNeXt-101 was deployed as the facial embedding generator and six face identification experiments were conducted.
	For each of the six attributes, Cumulative Match Curves (CMC) are provided in Fig.~\ref{FIG:DRFR-CelebA-CMC} w/ and w/o that attribute, respectively.
	Note that since there are different identities and a different number of images involved in each experiment, the horizontal comparison of the identification rate across attributes does not lead to meaningful conclusions.
	
	In Fig.~\ref{FIG:DRFR-CelebA-CMC}, the identification rates with and without each attribute are presented. Our results indicate that the face recognition performance decreases in the presence of the attributes \texttt{Bangs}, \texttt{Eyeglasses}, \texttt{Mustache}, \texttt{Sideburns}, and \texttt{Wearing Hat}.
	The attributes can be ranked according to the rank-1 identification rate degradation as follows: \texttt{Eyeglasses} (18.23\%) $>$ \texttt{Wearing Hat} (12.14\%) $>$ \texttt{Bangs} (6.97\%) $>$ \texttt{Sideburns} (2.56\%) $\sim$ \texttt{Mustache} (2.41\%).
	These results demonstrate that occlusion originating from facial accessories (\ie, eyeglasses, and hat) as well as facial hair (\ie, mustache, bangs, and sideburns) is an important challenge that affects the performance of face recognition algorithms. 
	Additionally, we observed that occlusion due to external accessories affects the performance more than occlusion originating from facial hair. 
	Finally, note that the identification performance is almost the same regardless of whether the subject is smiling or not. 
	The main reason for these results is that such datasets are collected from the web and thus, they usually cover a large range of head poses and contain enough images of smiling individuals. However, there is still a \emph{high imbalance} in other factors such as occlusion, which reduces the robustness of face recognition methods. While testing the facial attribute predictor on the VGGFace2 dataset we observed class imbalance ranging from 19:1 (\texttt{Bangs}) to 6:1 (\texttt{Wearing Hat}) in the VGGFace2 dataset.
	
	\section{Improving the Generalization}
	\label{SEC:TPL_Gem}
	
	The training process of \oreo (depicted in Fig.~\ref{FIG:DRFR-Framework}\!~(a)) consists of 
	(i) an occlusion-balanced sampling (OBS) to address the occlusion imbalance; 
	(ii) an occlusion-aware attention network (OAN) to jointly learn the global and local features;
	and (iii) the objective functions that guide the training process.
	Aiming to balance the occluded and non-occluded images within the batch, random pairs of non-occluded and occluded images are sampled and provided as input to the network. 
	Then, the proposed attention mechanism is plugged into the backbone architecture to generate the attention mask and aggregate the local with the global features to construct a single template.
	The final aggregated features are trained to learn occlusion-robust template guided by the softmax cross-entropy, sigmoid cross-entropy, and similarity triplet loss (STL) functions.
	
 	\noindent \textbf{Occlusion-Balanced Sampling}.
	In a hypothetical scenario in which the training data would be accompanied by occlusion ground-truth labels, the training set could easily be split into two groups of occluded and non-occluded images from which balanced batches could be sampled and fed to the neural network. However, this is not the case with existing face recognition training datasets since occlusion ground-truth labels are not provided. Aiming to generate occlusion labels, we focused on facial attributes that contain occlusion information. A state-of-the-art face attribute predictor~\cite{Sarafianos_2018_180921} was trained on the Celeb-A dataset and it was then applied to the training set to generate pseudo-labels. Those attribute pseudo-labels can then be utilized to facilitate occlusion-balanced sampling during training.
	It is worth noting this approach is dataset-independent and can be generalized to any face dataset since we solely rely on the attribute predictions learned from the Celeb-A dataset.
	By using this strategy, the network $\tplgen$ is feed-forwarded with pairs of randomly chosen non-occluded and occluded images denoted by $\{\{\im_n, \gtlabel_n\}, \{\im_o, \gtlabel_o\}\}_i, i=\{1, \ldots, N\}$, where $\gtlabel$ contains the identity and attributes of the facial images and $N$ is the total number of pairs. 
	Since OBS randomly generates pairs of occluded and non-occluded facial images in each epoch, their distribution within the batch is ensured to be balanced. 
	
	%\subsection{Occlusion-aware Attention Network}
	\noindent \textbf{Occlusion-aware Attention Network}.
	The template generator $\tplgen$ consists of three components: (i) a backbone network $\backboneblock$, (ii) an attention mechanism $\attblock$, and (iii) a feature aggregation module $\fuseblock$.
	Features are generated from two pathways as depicted in Figure~\ref{FIG:DRFR-Framework}\!~(b): a bottom-up for the global representations and a top-down for the local features. The bottom-up term describes the process of using representations of larger spatial dimensions to obtain a low-dimensional global embedding. The top-down term uses low-dimensional feature embeddings as input to the proposed attention masks applied in higher dimensions to generate local representations. 
	In the bottom-up pathway, the process of the network to generate global features is described by $\globaltpl = \backboneblock(\im)$.
	%In this paper, a ResNeXt-101~\cite{He_2016_190212} network is used as a backbone architecture. 
	In the top-down pathway, since the global features include information from the occluded region of the face, an attention module $\attblock$ is proposed that distills the identity-related features from the global feature maps to the local representations $\localtpl$.
	Finally, a feature aggregation module $\fuseblock$ is employed that aggregates the global and local features into a single compact representation $\tpl$.
	The objective of attention modules $\attblock$ is to help the model identify which areas of the original image contain important information based on the identity and attribute labels. 
	Assuming the feature maps extracted from different blocks of the backbone network are denoted by $\{B_1, B_2, B_3, B_4\}$, then the two-step attention mechanism is designed as follows.
	In the first level (denoted by \(\att_3\) in Fig.~\ref{FIG:DRFR-Framework}\!~(b)), we focus on self-attention and thus, the feature map $B_3$ is first broadcasted and then added with the global representation $\globaltpl$ to generate the attention mask $\att_{3}$. 
	The objective of $\att_{3}$ is to find the high-response region of the feature map by giving emphasis to the identity-related features and construct the local representation $\tpl^{l_3}$.
	This process is described by the following equation:
	\begin{equation}
	\label{EQU:DRFR-ATT3}
	\tpl^{l}_3 = \att_{3} * B_3 = h_3(\globaltpl, B_3) * B_3\ ,
	\end{equation}
	where $h_3$ is a sequence of convolution layers to reduce the channels and generate a single-channel attention map and the ``$*$'' operation corresponds to element-wise multiplication.
	The final global feature $\globaltpl$ is preserved as part of the final representation of the network so that $\globaltpl$ learns identity-related information.
	Thus, $\globaltpl$ guides the network to learn local attention maps on features from the previous block and distill the identity information from the most discriminative region to construct $\localtpl$. 
Our experiments indicated that the most discriminative region on the face from our first level attention mechanism is the eye region. While this is sufficient for most attributes, this is not the case when the individual is wearing glasses.
	To improve the generalization ability of the model in such cases, an additional attention mechanism is introduced on the feature map $B_2$ to force the network to focus on other regions of the face. 
	In the second level (denoted by $A_2$ in Fig.~\ref{FIG:DRFR-Framework}~(b)), the attention map is guided by the facial attribute predictions in a weakly-supervised manner (\ie, no attribute ground-truth information is utilized since we only have access to the visual attribute predictions). 
	Thus, the local representations at this level are computed by:
	\begin{equation}
	\label{EQU:DRFR-ATT4}
	\tpl^{l}_2 = (1 - \att_{2}) * B_2 = (1 - h_2(\globaltpl,  B_2)) * B_2\ ,
	\end{equation}
	where $h_2$ is an attention operation guided by both identity labels and attributes labels. 
	Since the attention map $A_2$ is guided not only by the identity loss function but also by the attribute predictions, the network is capable of focusing on image regions related to both the identity and the visual attributes. 
	The global and local features $\{\globaltpl, \tpl^{l}_2, \tpl^{l}_3\}$ are concatenated into one single vector and are projected in a single feature template $\tpl$, which enforces both global and local features to preserve semantic identity information. 
	
	\begin{table*}[thb]
		\begin{center}
			\begin{adjustbox}{max width=\linewidth}
				\begin{tabular}{lcccccccccccc}
					\toprule
					\multirow{2}{*}{Method} & \multicolumn{2}{c}{\texttt{Bangs}} & \multicolumn{2}{c}{\texttt{Eyeglasses}} & \multicolumn{2}{c}{\texttt{Mustache}} & \multicolumn{2}{c}{\texttt{Sideburns}} & \multicolumn{2}{c}{\texttt{Wearing Hat}} & \multirow{2}{*}{ADP (\%)} \\ \cmidrule(r){2-3} \cmidrule(r){4-5} \cmidrule(r){6-7} \cmidrule(r){8-9} \cmidrule(r){10-11}
					& w/o & w/ & w/o & w/ & w/o & w/ & w/o & w/ & w/o & w/\\
					\midrule 
					ResNeXt-101~\cite{He_2016_190212} & $73.76$ & $66.79$ & $81.38$ & $63.14$ & $83.26$ & $80.85$ & $81.47$ & $78.91$ & $77.74$ & $65.60$ & $8.46$\\
					%\midrule
					\oreo & $\mathbf{74.65^{\ast}}$ & $\mathbf{68.66^{\ast}}$ & $\mathbf{81.99}$ & $\mathbf{65.53^{\ast}}$ & $\mathbf{84.72^{\ast}}$ & $\mathbf{82.53^{\ast}}$ & $\mathbf{83.13^{\ast}}$ & $\mathbf{81.47^{\ast}}$ & $\mathbf{80.02^{\ast}}$ & $\mathbf{68.34^{\ast}}$ & $\mathbf{7.60}$\\
					%\oreo & $\mathbf{75.03}$ & $\mathbf{69.18}$ & $\mathbf{82.11}$ & $\mathbf{65.75}$ & $\mathbf{85.45}$ & $\mathbf{82.09}$ & $\mathbf{83.24}$ & $\mathbf{81.02}$ & $\mathbf{80.36}$ & $\mathbf{69.02}$ & $\mathbf{7.83}$\\
					\bottomrule
				\end{tabular}
			\end{adjustbox}
		\end{center}
		\vspace{-1.7em}
		\caption{Comparison of rank-1 identification rate (\%) on the Celeb-A dataset w/ and w/o the specified attribute. ADP corresponds to the average degradation percentage. The lower the value of ADP, the more robust the method is. Note that ``${\ast}$'' denotes statistically significant improvements using the McNemar's statistical test.}
		\vspace{-1em}
		\label{TAB:DRFR-EXP-Celeb}
	\end{table*}
	
	\noindent \textbf{Learning Objectives}.

	Loss functions are employed for training: (i) the softmax cross-entropy loss $L_C$ for identity classification, (ii) the sigmoid binary cross-entropy loss $L_A$ for attribute prediction, and (iii) a new loss $L_T$ designed for the occlusion-balanced sampling. 
	The identity classification loss is defined as:
	\begin{equation}
	L_C = - \frac{1}{\bs} \sum_{i=0}^{\bs} log \frac{\exp\left(W_{y_i^c}t_i + b_{y_i^c}\right)}{\sum_{j=1}^{n} \exp\left(W_{j}t_{i} + b_j\right)},
	\end{equation}
	where $t_i$ and $y_i^{c}$ represent the features and the ground-truth identity labels of the $i^{th}$ sample in the batch, $W$ and $b$ denote the weights and bias in the classifier, and $\bs$ and $n$ correspond to the batch size and the number of identities in the training set, respectively.
	Following that, the sigmoid binary cross-entropy loss $L_A$ can be defined as
	\begin{equation}
	L_A = - \frac{1}{\bs} \sum_{i=0}^{\bs} \log \sigma(\attr_i) y^{\attr}_i + \log (1 - \sigma(\attr_i)) (1 - y^{\attr}_i)\ ,
	\end{equation}
	where $y^{\attr}$ corresponds to the attribute labels (pseudo-labels obtained by the attribute predictor) and $\sigma(\cdot)$ is the sigmoid activation applied on the attribute predictions $\attr$ (i.e., predictions of the occlusion-aware face recognition network $G$).
	
	In the matching stage, the cosine distance is used to compute the similarity between two feature embeddings.
	Since images with the same identity have a higher similarity score than those with a different identity, we introduce a similarity triplet loss (STL) as regularization \cite{Xu_2019_190315} to make the final facial embedding more discriminative.
	During training, each batch comprises pairs of non-occluded and occluded images with each pair having the same identity. 
	Let $\tpl_n$ and $\tpl_o$ be the final feature representations of non-occluded images $\im_n$ and occluded images $\im_o$ respectively. The similarity matrix $\simmtx \in \mathbb{R}^{\bs \times \bs}$ within the batch is then computed, where $\bs$ is the batch size.
	In the similarity matrix $\simmtx$, we aim to identify:
	(i) hard positives which are pairs of samples that originate from the same identity but have low similarity score $\simscore^p(\tpl_n, \tpl_o)$ and (ii) hard negatives which are pairs of samples with different identities but with high similarity score $\simscore^n(\tpl_n, \tpl_o)$. 
	Then, the objective function is defined as: 
	\begin{equation}
	L_{T} = \sum_{i=1}^{\bs} [\simscore^n_i(\tpl_n, \tpl_o) - \simscore^p_i(\tpl_n, \tpl_o) + \margin]_{+}\ .
	\end{equation}
	A margin $\margin \in \mathbb{R}^{+}$ is maintained to enforce that small angles (high similarity score) belong to the same identity and large angles (low similarity score) belong to different identities. 
	Finally, the whole network is trained using the summation of the individual losses: \(L = L_{C} + L_{A} + L_{T}\).
	
	\begin{figure*}[tb]
		\centering
		\includegraphics[width=\linewidth]{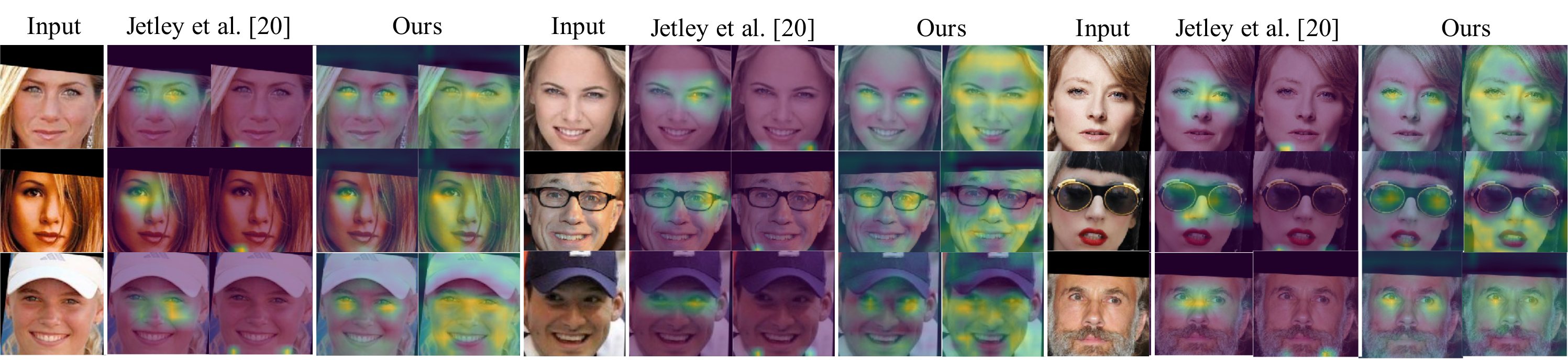}
		\vspace{-2em}
		\caption{Visualization of the discriminative region using an attention mechanism. The first attention map of \oreo focuses on the discriminative region (eyes), whereas the second attention map focuses on the non-occluded facial region.}
		\vspace{-0.5em}
		\label{FIG:DRFR-CFP-ATT}
	\end{figure*}
	
	\section{Experiments}
	Four evaluation datasets are used to extensively evaluate our algorithm under different scenarios. 
	To generate the occlusion meta-data (\ie, attributes) of the training set, a state-of-the-art face attribute predictor \cite{Sarafianos_2018_180921} is used to predict the occlusion-related attributes on the training set. 
	The evaluation protocol of each dataset is strictly followed. For face verification, Identification Error Trade-off is reported with true acceptance rates (TAR) at different false accept rates (FAR). For face identification, CMC is reported. These evaluation metrics were computed using FaRE toolkit \cite{Xu_2019_190129}. 
	
	\begin{table*}[thb]
	    \begin{center}
	        \begin{adjustbox}{max width=\linewidth}
	            \begin{tabular}{c | c  c  c  c  c  c  c}
	                \toprule
	                \diagbox{Pitch}{Yaw} & $-90^{\circ}$ & $-60^{\circ}$ &  $-30^{\circ}$ & $0^{\circ}$ & $+30^{\circ}$ & $+60^{\circ}$ & $+90^{\circ}$ \\
	                \midrule
	                $+30^{\circ}$ &
	                \begin{tabular}[c]{@{}c@{}}$58$, $82$, $\mathbf{96}$\end{tabular} & 
	                \begin{tabular}[c]{@{}c@{}}$95$, $99$, $\mathbf{100}$\end{tabular} & 
	                \begin{tabular}[c]{@{}c@{}}$100$, $100$, $\mathbf{100}$\end{tabular} & 
	                \begin{tabular}[c]{@{}c@{}}$100$, $100$, $\mathbf{100}$\end{tabular} & 
	                \begin{tabular}[c]{@{}c@{}}$99$, $99$, $\mathbf{100}$\end{tabular}& 
	                \begin{tabular}[c]{@{}c@{}}$92$, $99$, $\mathbf{100}$\end{tabular}& 
	                \begin{tabular}[c]{@{}c@{}}$60$, $75$, $\mathbf{94}$\end{tabular}\\
	                
	                $0^{\circ}$ & 
	                \begin{tabular}[c]{@{}c@{}}$84$, $96$, $\mathbf{98}$ \end{tabular}& 
	                \begin{tabular}[c]{@{}c@{}}$99$, $100$, $\mathbf{100}$\end{tabular}& 
	                \begin{tabular}[c]{@{}c@{}}$100$, $100$, $\mathbf{100}$\end{tabular}& 
	                - & 
	                \begin{tabular}[c]{@{}c@{}}$100$, $100$, $\mathbf{100}$\end{tabular}& 
	                \begin{tabular}[c]{@{}c@{}}$99$, $100$, $\mathbf{100}$\end{tabular} & 
	                \begin{tabular}[c]{@{}c@{}}$91$, $96$, $\mathbf{100}$\end{tabular} \\
	                
	                $-30^{\circ}$ & 
	                \begin{tabular}[c]{@{}c@{}}$44$, $74$, $\mathbf{86}$\end{tabular}& 
	                \begin{tabular}[c]{@{}c@{}}$80$, $97$, $\mathbf{99}$\end{tabular}& 
	                \begin{tabular}[c]{@{}c@{}}$99$, $100$, $\mathbf{100}$\end{tabular}& 
	                \begin{tabular}[c]{@{}c@{}}$99$, $100$, $\mathbf{100}$\end{tabular}& 
	                \begin{tabular}[c]{@{}c@{}}$97$, $100$, $\mathbf{100}$\end{tabular}& 
	                \begin{tabular}[c]{@{}c@{}}$90$, $96$, $\mathbf{100}$\end{tabular}& 
	                \begin{tabular}[c]{@{}c@{}}$35$, $78$, $\mathbf{95}$\end{tabular}\\
	                \bottomrule
	            \end{tabular}
	        \end{adjustbox}
	    \end{center}
	    \vspace{-1.3em}
	    \caption{Comparison of Rank-1 identification rate (\%) of different methods on the \texttt{UHDB31.R128.I03} dataset. The results in each cell are ordered in the sequence of FaceNet \cite{Schroff_2015_180917}, UR2D-E \cite{Xu_2017_17643}, and \oreo.}
	    \label{TAB:DRFR-EXP-UHDB31}
	   \vspace{-0.5em}
	\end{table*}
	
	\subsection{Comparison with state of the art}
	\noindent \textbf{Celeb-A: In-the-wild Face Identification.}
	We tested our algorithm on the Celeb-A dataset~\cite{Liu_2015_180921} under various types of occlusion. It is worth noting that the identity labels of the Celeb-A dataset were not utilized since \oreo was trained solely on the VGGFace2 dataset.     
	The rank-1 identification rate w/ and w/o each attribute is presented in Table~\ref{TAB:DRFR-EXP-Celeb}. 
	We observe that \oreo outperforms ResNeXt-101 in all settings (w/ and w/o attributes), which suggests that our algorithm can learn robust discriminative feature representations regardless of occlusions. The improvement is statistically significant in all attributes according to the McNemar's test. In addition, \oreo demonstrated a lower average degradation percentage than the baseline by $10.17\%$ in terms of relative performance. 
	This indicates that our algorithm improves the generalization ability of the facial embedding generator in the presence of occlusions.
	
	\begin{table}[tb]
		\begin{center}
			\begin{adjustbox}{max width=\linewidth}
				\begin{tabular}{lcccc}
					\toprule
					\multirow{2}{*}{Method} & \multirow{2}{*}{Acc. (\%)}
					& \multicolumn{3}{c}{TAR (\%) @ FAR=} \\ \cmidrule{3-5} 
					&   & $10^{-3}$ & $10^{-2}$ & $10^{-1}$ \\ 
					\midrule
					DR-GAN~\cite{Tran_2017_17705} & $93.4 \pm 1.2$ & - & - & - \\
					MTL-CNN~\cite{Yin_2018_181109} & $94.4 \pm 1.2$ & - & - & - \\
					ArcFace~\cite{Deng_2019_180917} & $93.9 \pm 0.8$ & $80.2 \pm 5.9 $ & $86.0 \pm 2.8$ & $94.3 \pm 1.5$ \\
					ResNeXt-101~\cite{He_2016_190212} & $97.1 \pm 0.8$ & $81.9 \pm 11.4$ & $92.3 \pm 4.1$ & $98.9 \pm 0.8$\\
					\midrule
					\oreo & $\mathbf{97.5} \pm \mathbf{0.5}$ & $\mathbf{85.5} \pm \mathbf{5.3}$ & $\mathbf{94.1} \pm \mathbf{2.5}$ & $\mathbf{99.2} \pm \mathbf{0.7}$\\
					\bottomrule
				\end{tabular}
			\end{adjustbox}
		\end{center}
		\vspace{-1.5em}
		\caption{Comparison of the face verification performance with state-of-the-art face recognition techniques on the CFP dataset using CFP-FP protocol. The evaluation results are presented by the average $\pm$ standard deviation over $10$ folds.}
		\label{TAB:DRFR-EXP-CFP}
		\vspace{-1.5em}
	\end{table}
	
	\begin{table*}[htb]
		\begin{center}
			\begin{adjustbox}{max width=\linewidth}
				\begin{tabular}{lccccccccccccc}
					\toprule
					\multirow{3}{*}{Method} & \multicolumn{7}{c}{1:1 Mixed Verification} & \multicolumn{5}{c}{1:N Mixed Identification} \\
					& \multicolumn{7}{c}{ TAR (\%) @ FAR=} & \multicolumn{3}{c}{TPIR (\%) @ FPIR=} & \multicolumn{3}{c}{Retrieval Rate (\%)}\\ \cmidrule(r){2-8} \cmidrule(r){9-11} \cmidrule(r){12-14}
					& $10^{-7}$ & $10^{-6}$ & $10^{-5}$ & $10^{-4}$ & $10^{-3}$ & $10^{-2}$ & $10^{-1}$ & $10^{-3}$ & $10^{-2}$ & $10^{-1}$ & Rank-1 & Rank-5 & Rank-10\\ \midrule
					GOTS~\cite{Maze_2018_180930} & $3.00$ & $3.00$ & $6.61$ & $14.67$ & $33.04$ & $61.99$ & $80.93$ & $2.66$ & $5.78$ & $15.60$ & $37.85$ & $52.50$ & $60.24$\\
					FaceNet~\cite{Schroff_2015_180917} & $15.00$ & $20.95$ & $33.30$ & $48.69$ & $66.45$ & $81.76$ & $92.45$ & $20.58$ & $32.40$ & $50.98$ & $69.22$ & $79.00$ & $81.36$ \\
					VGGFace~\cite{Parkhi_2015_16638} & $20.00$ & $32.20$ & $43.69$ & $59.75$ & $74.79$ & $87.13$ & $95.64$ & $26.18$ & $45.06$ & $62.75$ & $78.60$ & $86.00$ & $89.20$\\
					MN-vc~\cite{Xie_2018_181002} & - & - & - & \textcolor{Blue}{\bm{$86.20$}} & \textcolor{Blue}{\bm{$92.70$}} & \textcolor{Blue}{\bm{$96.80$}} & \textcolor{Blue}{\bm{$98.90$}} & - & - & - & - & - & -\\
					ArcFace~\cite{Deng_2019_180917} & $\mathbf{60.50}$ & $\mathbf{73.56}$ & $\mathbf{81.70}$ & $\mathbf{87.90}$ & $91.14$ & $95.98$ & $97.92$ & $\mathbf{70.90}$ & $\mathbf{81.98}$ & $\mathbf{87.63}$ & \textcolor{Blue}{\bm{$92.25$}} & \textcolor{Blue}{\bm{$94.31$}} & \textcolor{Blue}{\bm{$95.30$}}\\
					\midrule
					ResNeXt-101~\cite{He_2016_190212} & $28.72$ & $58.09$ & $71.19$ & $81.76$ & $90.70$ & $95.75$ & $98.86$ & $53.66$ & $71.50$ & $82.47$ & $91.88$ & $95.51$ & $97.29$\\
					\oreo  & \textcolor{Blue}{\bm{$51.97$}} & \textcolor{Blue}{\bm{$62.36$}} & \textcolor{Blue}{\bm{$75.86$}} & $85.19$ & $\mathbf{92.81}$ & $\mathbf{97.11}$ & $\mathbf{99.37}$& \textcolor{Blue}{\bm{$65.47$}} & \textcolor{Blue}{\bm{$77.11$}} & \textcolor{Blue}{\bm{$85.92$}} & $\mathbf{93.76}$& $\mathbf{96.68}$ & $\mathbf{97.74}$\\
					\bottomrule
				\end{tabular}
			\end{adjustbox}
		\end{center}
		\vspace{-1.4em}
		\caption{Comparison of the face verification and identification performance of different methods on the IJB-C dataset. Top performance is marked with \textbf{black} and second-best with \textcolor{Blue}{\textbf{blue}}.
		\oreo significantly improves the face verification and identification performance compared to the baseline, and achieves state-of-the-art results in terms of retrieval rate.}
		%\vspace{-0.5em}
		\label{TAB:DRFR-EXP-IJBC}
	\end{table*}
	
	\begin{table*}[htb]
		\begin{center}
			\begin{adjustbox}{max width=\linewidth}
				\begin{tabular}{lccccccccccccccc}
					\toprule
					\multirow{3}{*}{Method} & \multicolumn{3}{c}{Module} & \multicolumn{4}{c}{CFP-FF} &  \multicolumn{4}{c}{CFP-FP} &   \multicolumn{4}{c}{LFW} \\
					\cmidrule(r){2-4} \cmidrule(r){5-8} \cmidrule(r){9-12}  \cmidrule(r){13-16}  
					& \multirow{2}{*}{OAN} & \multirow{2}{*}{OBS} & \multirow{2}{*}{STL} & \multirow{2}{*}{Acc. (\%)} & \multicolumn{3}{c}{TAR (\%) @ FAR=} & \multirow{2}{*}{Acc. (\%)} &  \multicolumn{3}{c}{TAR (\%) @ FAR=}& \multirow{2}{*}{Acc. (\%)} &  \multicolumn{3}{c}{TAR (\%) @ FAR=}\\ \cmidrule(r){6-8} \cmidrule(r){10-12} \cmidrule(r){14-16}
					& & & & & $10^{-3}$ & $10^{-2}$ & $10^{-1}$ & & $10^{-3}$ & $10^{-2}$ & $10^{-1}$ & & $10^{-3}$ & $10^{-2}$ & $10^{-1}$ \\
					\midrule
					ResNeXt-101~\cite{He_2016_190212} & & & & $94.77$ & $43.71$& $82.15$& $97.03$& $88.66$ & $32.71$ &$50.20$ & $85.17$ &  $94.77$ & $64.77$ & $84.50$ & $96.31$ \\
					Jetley \etal~\cite{Jetley_2018_181010} & \checkmark & & & $93.47$ & $40.48$& $78.07$& $94.75$& $83.71$ & $22.23$ &$39.07$ & $71.31$ &  $92.70$ & $55.63$ & $77.23$ & $93.61$ \\
					\midrule
					\multirow{4}{*}{\oreo} & \checkmark & & & $95.41$& $49.69$ & $85.68$ & $\mathbf{97.56}$ & $90.24$ & $35.34$ & $59.87$ & $89.33$ & $95.90$& $66.33$ & $90.00$ & $97.47$  \\
					& & \checkmark & & $95.33$ & $46.02$ & $86.01$ & $97.53$ & $89.21$ & $33.49$ & $56.51$ & $87.49$ & $95.55$ & $71.63$ & $89.80$ & $97.12$ \\
					& & \checkmark  & \checkmark & $95.66$ & $48.39$ & $87.76$ & $97.74$ & $89.49$ & $36.97$ & $59.56$ & $88.31$ & $96.17$ & $\mathbf{74.13}$ & $90.63$ & $97.63$ \\
					\cmidrule(r){2-16}
					& \checkmark & \checkmark  & \checkmark & $\mathbf{95.86}$ & $\mathbf{49.74}$ & $\mathbf{89.38}$ & $97.48$ & $\mathbf{90.60}$ & $\mathbf{40.06}$ & $\mathbf{63.90}$ & $\mathbf{90.64}$ & $\mathbf{96.20}$ & $68.13$ & $\mathbf{92.03}$ & $\mathbf{97.86}$ \\
					\midrule
				\end{tabular}
			\end{adjustbox}
		\end{center}
		\vspace{-1.5em}
		\caption{Impact of the individual proposed components under face verification protocols on the CFP and LFW datasets.}
		\vspace{-1em}
		\label{TAB:DRFR-AB-Study}
	\end{table*}
	
	\noindent \textbf{CFP: In-the-wild Face Verification.}
	The CFP~\cite{Sengupta_2016_17835} is used to evaluate face verification performance on images in-the-wild, which contain variations in pose, occlusion, and age. 
	In the first experiment, we qualitatively visualize the region of frontal-face images that demonstrated the highest response, because such regions provide the most meaningful information to the final embedding.
	We chose a state-of-the-art algorithm~\cite{Jetley_2018_181010} as our baseline to compare the attention mask learned from the data.
	Figure~\ref{FIG:DRFR-CFP-ATT} depicts samples from the CFP dataset with different occlusion variations.
	Compared to the attention mask of Jetley \etal~\cite{Jetley_2018_181010}, our attention mechanism has the following advantages: 
	(i) By combining the global with the local feature maps to generate the attention mask, the global representation demonstrates a higher response around the eye regions, which indicates that the eyes contain discriminative identity-related information.
	It also explains why the \texttt{Eyeglasses} attribute demonstrated the highest performance drop in Section~\ref{SEC:DRFR-CelebA};
	(ii) By learning the attention from local feature maps guided by the occlusion pseudo-labels generated by our method, we observe that the proposed attention mask is focusing more on the non-occluded region.
	In addition, it helps the embeddings aggregate information from the non-occluded facial region instead of solely relying on the eye regions as indicated by our first observation.
	Note that, the self-attention described in Eq.~(\ref{EQU:DRFR-ATT3}) is learned directly from the data without having access to ground-truth masks that would help our model identify better the most discriminative regions (hence the failure case depicted in the second to last image of the second row in Figure~\ref{FIG:DRFR-CFP-ATT}). This is why an additional attention map was introduced in Eq.~(\ref{EQU:DRFR-ATT4}) which is learned in a weakly-supervised manner from the attribute predictions and helps our algorithm to focus on the non-occluded region (right-most image of the second row).
	
	In the second experiment, we used the CFP-FP protocol to quantitatively evaluate the face verification performance and the experimental results are presented in Table~\ref{TAB:DRFR-EXP-CFP}.
	We used the following metrics to evaluate the performance: (i) the verification accuracy, and (ii) the TAR at FAR equal to $10^{-3}$, $10^{-2}$, and $10^{-1}$.
	Compared to all baselines, \oreo achieves state-of-the-art results on this dataset in terms of accuracy and increases the TAR at low FARs.
	The moderately better accuracy results demonstrate that \oreo can also improve the performance of general face recognition.
	
	\noindent \textbf{UHDB31: Lab-controlled Face Identification.} The UHDB31~\cite{Le_2017_180921} dataset is used to evaluate the face identification performance under different pose variations. \oreo outperformed FaceNet~\cite{Schroff_2015_180917} and UR2D-E \cite{Xu_2017_17643} in all settings and especially in large pose variations (\eg, yaw = \(-90^o\), pitch = \(+30^o\), which indicates that our algorithm is robust to pose variations (Table~\ref{TAB:DRFR-EXP-UHDB31}). 
	
	\noindent \textbf{IJB-C: Set-based Face Identification and Verification.}
	The IJB-C dataset~\cite{Maze_2018_180930} is a mixed media set-based dataset with open-set protocols comprising images with different occlusion variations.  
	Two experiments are performed on this dataset following 1:1 mixed verification protocol and 1:N mixed identification protocol.
	To generate the facial embedding for the whole set, the corresponding images are fed to the neural networks and the average of the embeddings from all images within a set is computed.
	The evaluation metrics include the verification metric of ROC, the identification metric of retrieval rate, the true positive identification rate (TPIR) at different false positive identification rates (FPIR). From the obtained results in Table~\ref{TAB:DRFR-EXP-IJBC}, we observe that \oreo outperforms four out of six baselines in all metrics and comes second to ArcFace only in some cases under the mixed verification and identification protocols. 
	ArcFace was trained on the MS1M~\cite{Guo_2016_180925} dataset which contains significantly more identities and data than the VGGFace2 dataset. 
	\oreo performs well across the board and outperforms ArcFace at high FARs in the verification protocol as well as in the identification retrieval protocol. When compared against the baseline, \oreo significantly improves the performance in both verification and identification. For example, when FAR is equal to $10^{-7}$ the TAR improves from 28.72\% to 51.97\%. These results demonstrate that \oreo can successfully learn features that are robust to occlusions. 
	
	\subsection{Ablation Study}
	An ablation study was conducted to explore the impact of the proposed components: (i) the attention mechanism (OAN), (ii) the balanced training strategy (OBS), and (iii) the similarity triplet loss (STL). The results of the contributions of each component along with the backbone network~\cite{He_2016_190212} and an attention-based baseline~\cite{Jetley_2018_181010} on the LFW~\cite{Huang_2007_180918} and CFP~\cite{Sengupta_2016_17835} datasets are depicted in Table~\ref{TAB:DRFR-AB-Study}. For faster turnaround, in this study we used only the first 500 subjects of VGGFace2~\cite{Cao_2018_17830} as training set which is why the results between Table~\ref{TAB:DRFR-EXP-CFP} and Table~\ref{TAB:DRFR-AB-Study} are different.
	Compared to the backbone network, \oreo increases the verification accuracy by at least $1.10\%$. 
	
	\noindent \textbf{Occlusion-aware Attention Network.}
	OAN helps the network focus on the local discriminative regions by pooling local features.
	From the results in Table~\ref{TAB:DRFR-AB-Study}, we observe that the backbone architecture that encodes only global representations achieves better results than the method of Jetley \etal~\cite{Jetley_2018_181010}, which leverages only local information. The proposed attention mechanism outperforms both techniques since it includes both global and local features into the final embedding guided by the identity loss.
	Finally, compared to ResNeXt-101 which served as our baseline, OAN results in absolute improvements ranging from \(0.64\%\) to \(1.58\%\) in terms of verification accuracy. 
	
	\noindent \textbf{Occlusion Balanced Sampling.} 
	OBS creates pairs of occluded and non-occluded images to alleviate the problem of occlusion imbalance during training. OBS results in absolute improvements ranging from \(0.56\%\) to \(0.78\%\) in terms of verification accuracy compared to the performance of the backbone network. These results indicate that OBS has a limited effect on these datasets because they contain a limited number of occluded samples.
	%\vspace{0.3cm}
	
	\noindent \textbf{Similarity Triplet Loss.}
	STL works as regularization to the final loss, as it increases the similarity of two feature embeddings that belong to the same identity and penalizes the network when the similarity is high but the features are originating from different identities. STL does not force the representations to be learned in the same scale which is an advantage compared to other alternative similarity loss functions~\cite{Deng_2019_180917, Liu_2017_180923, Wang_2018_181025}.
	Since STL requires pairs of occluded and non-occluded images, OBS is a prerequisite for this loss, which is why an experimental result with only STL is not provided in Table~\ref{TAB:DRFR-AB-Study}.
	We observe that STL improves the performance by at least $0.83\%$ in terms of verification accuracy. 
	In addition, by comparing the TARs at low FARs on all three datasets, we observe that OBS along with STL are the components that affect the most performance compared to ResNeXt-101. 
	The obtained results demonstrate that the features learned with STL are more discriminative than those learned using only a softmax cross-entropy loss.
	
	\section{Conclusion}
	In this paper, we systematically analyzed the impact of facial attributes to the performance of a state-of-the-art face recognition method and quantitatively evaluated the performance degradation under different types of occlusion caused by facial attributes. 
	To address this degradation, we proposed \oreo: an occlusion-aware approach that improves the generalization ability on facial images occluded. 
	An attention mechanism was designed that extracts local identity-related information.  In addition, a simple yet effective occlusion-balanced sampling strategy and a similarity-based triplet loss function were proposed to balance the non-occluded and occluded images and learn more discriminative representations.
	Through ablation studies and extensive experiments, we demonstrated that \oreo achieves state-of-the-art results on several publicly available datasets and provided an effective way to better understand the representations learned by the proposed method.
	
	\smallskip
	\small{\noindent \textbf{Acknowledgment} This material was supported by the U.S. Department of Homeland Security under Grant Award Number 2017-STBTI-0001-0201 with resources provided by the Core facility for Advanced Computing and Data Science at the University of Houston.}
	
	{\small
		\bibliographystyle{ieee_fullname}
		\bibliography{egbib}
	}
	
\end{document}